\documentclass[conference]{IEEEtran}
\IEEEoverridecommandlockouts
\usepackage{cite}
\usepackage{amsmath,amssymb,amsfonts}
\usepackage{algorithmic}
\usepackage{graphicx}
\usepackage{textcomp}
\usepackage{xcolor}
\usepackage{orcidlink}
\usepackage{tabularx}  
\usepackage{booktabs}
\usepackage{multirow}
\def\BibTeX{{\rm B\kern-.05em{\sc i\kern-.025em b}\kern-.08em
 T\kern-.1667em\lower.7ex\hbox{E}\kern-.125emX}}
\begin{document}

\title{Surgical Video Generation From
Diffusion to World Models: A Survey\
\thanks{This work was partially supported by National Natural Science Fund of China under Grants 92570110 and 62271090, Chongqing Natural Science Fund under Grant CSTB2024NSCQ-JQX0038, and National Youth Talent Project.}
}

\author{\IEEEauthorblockN{Fuxiang Huang\orcidlink{0000-0003-0399-9932}}
\IEEEauthorblockA{\textit{School of Data Science} \\
\textit{Lingnan University}\\
Hong Kong, China \\
fxhuang1995@gmail.com}
\and
\IEEEauthorblockN{Chenxu Zhang; Liang Han; Lei Zhang\textsuperscript{*}\orcidlink{0000-0002-5305-8543} }
\IEEEauthorblockA{\textit{School of Microelectronics and Communication Engineering} \\
\textit{Chongqing University}\\
Chongqing, China \\
zhangchenxu@cqu.edu.cn; hanliangaa@cqu.edu.cn; leizhang@cqu.edu.cn}
}

\maketitle

\begin{abstract}
Surgical video data provides the primary training resource for models of intraoperative perception, surgical workflow understanding, and robotic decision-making. However, clinical data acquisition remains constrained by privacy, cost, and class imbalance. Surgical video generation has emerged as a transformative approach to addressing data scarcity and as a foundation for surgical simulation, training, and robotic policy learning. The field has developed rapidly without a clear conceptual framework. This survey organizes the 2024--2026 literature into three categories: unconditional generation, conditional generation, and world modeling generation, revealing a fundamental shift in how the task is defined from synthesizing visually plausible frames to modeling the causal dynamics of surgical scenes. We examine the persistent gap between pixel-level fidelity and clinical plausibility, and identify generalization, physical realism, controllability, and interpretability as bottlenecks. We further summarize experimental results of representative methods on public datasets to provide a quantitative reference for the field. This survey provides a structured overview of the current state and open challenges, offering a reference for researchers working at the intersection of intelligent perception, multi-modal fusion, generative AI, and surgical data science.
\end{abstract}

\begin{IEEEkeywords}
surgical video generation, diffusion models, multi-modal fusion, world model, intelligent perception 
\end{IEEEkeywords}

\section{Introduction}
Artificial intelligence (AI) in surgery aims to bridge computational intelligence and clinical practice, delivering reproducible, safe, and scalable technologies for surgical training, decision support, and robotic automation \cite{maier2022surgical}. High-quality surgical videos constitute the primary data substrate for building computer vision models that parse surgical workflows, quantify surgeon behaviors, and optimize robotic manipulation strategies. However, real-world surgical video datasets face three fundamental practical limitations. First, privacy regulations hinder large-scale public data dissemination. Second, rare surgical cases and infrequent intraoperative adverse events lead to severe long-tailed data distributions. Third, manual labeling for surgical phases, instruments, and tissue-tool interactions is labor-intensive, costly, and prone to human error. These bottlenecks substantially obstruct the large-scale real-world deployment of surgical AI systems.

Generative AI has substantially advanced biomedical data engineering by enabling the synthesis of clinically meaningful visual content \cite{algethami2025generative}. Recent advances in video diffusion models \cite{ho2022video}, large vision-language models \cite{Chen_LLAMAVG_ISBI2025}, and neuro-symbolic systems \cite{Sivakumar_SWoMo_MICCAI2026}, in particular, have driven significant progress in controllable surgical video generation. Unlike generic video synthesis tasks, surgical video generation imposes stringent domain constraints: synthesized outputs must preserve correct anatomical topology, conform to standard surgical procedural logic, and guaranty physically plausible tissue-tool interactions. Such specialized requirements make surgical video generation a unique interdisciplinary research problem that requires a customized methodological investigation.

Driven by these demands, research on surgical video generation has grown rapidly from 2024 to 2026. Despite this rapidly expanding body of literature, existing contributions remain fragmented and lack a unified conceptual framework. To the best of our knowledge, no dedicated survey offers a domain-focused, critical review of this rapidly evolving research topic. Representative prior work such as Algethami \textit{et al.} \cite{algethami2025generative} reviews biomedical video synthesis from the perspective of imaging modalities, spanning endoscopy, echocardiography, MRI, and CT. Their work delivers a valuable cross-modal summary of general techniques and shared challenges across biomedical visual generation. Different from this modality-centered perspective, the present survey concentrates exclusively on surgical video generation. We structure existing literature by generation objective, tracing the technical trajectory from unconditional generation toward world modeling approaches. Concretely, we categorize existing approaches into three paradigms: \textit{unconditional generation}, \textit{conditional generation}, and \textit{world modeling generation}. This taxonomy demonstrates a clear technical evolution: research objectives shift from producing visually realistic individual frames toward modeling causal dynamics within surgical scenes. 
Furthermore, we systematically analyze prevailing evaluation protocols, highlight the persistent mismatch between pixel-level visual fidelity and clinical plausibility, and summarize core bottlenecks covering generalization capability, physical realism, controllability, and interpretability. Finally, we suggest a few directions that may merit further exploration.

The remainder of this survey is structured as follows. Section \ref{Problem Formulation} defines the task; Section \ref{Taxonomy of Surgical Video Generation} presents a taxonomy and reviews three paradigms; Section \ref{Datasets and Evaluation} covers datasets and metrics; Section \ref{Limitations and Future Directions} discusses limitations and outlines potential directions; Section \ref{Conclusion} concludes.

\begin{figure}[t]
 \centering \includegraphics[width=0.5\textwidth]{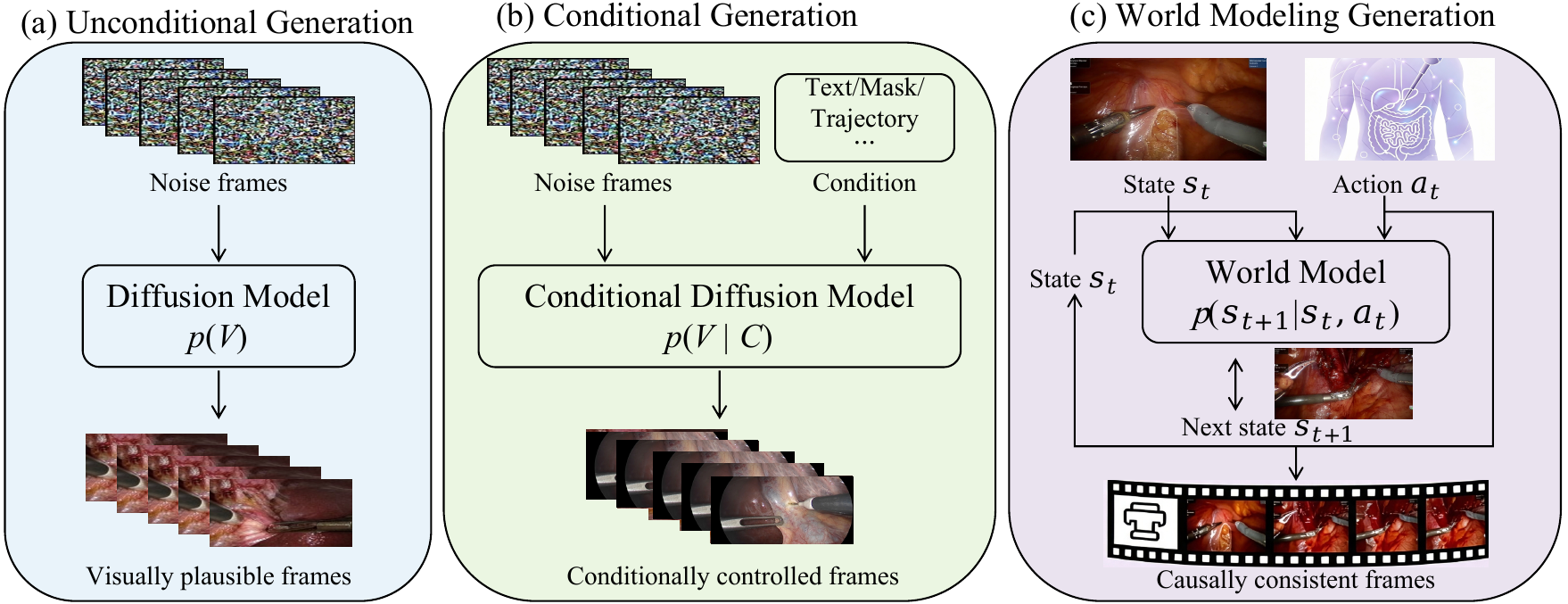}
 \caption{{Three paradigms of surgical video generation. (a) Unconditional Generation: visually plausible video generation from noise. (b) Conditional Generation: semantically controllable video generation guided by conditioning signals. (c) World Modeling Generation: causally consistent and physically plausible video generation through explicit state transition modeling.}
}
 \label{fig:paradigm_compare}
\end{figure}
\section{Problem Formulation}\label{Problem Formulation}
Let ${V} = \{v_1, \ldots, v_T\}$ denote a surgical video sequence of $T$ frames. Surgical video generation aims to learn a conditional distribution $p({V} \mid {C})$, where ${C}$ denotes control signals and may be empty. When present, ${C}$ provides control over the generation process. In conditional generation, ${C}$ specifies the desired content, such as texts, masks, or phase labels. In world modeling generation, ${C}$ represents actions that drive state transitions, shifting the focus from appearance control to modeling how surgical scenes evolve in response to interventions.
\section{Taxonomy of Surgical Video Generation}\label{Taxonomy of Surgical Video Generation}
We categorize existing methods according to their generation objective: \textit{unconditional generation}, \textit{conditional generation}, and \textit{world modeling generation}. 
Figure~\ref{fig:paradigm_compare} illustrates the distinction among the three paradigms.

\subsection{Unconditional Generation}
Unconditional generation, employs diffusion models to synthesize visually plausible surgical videos from random noise without external conditioning. Endora \cite{Li_Endora_MICCAI2024} pioneered a diffusion Transformer for endoscopic simulation, learning the unconditional distribution of surgical video data. This work demonstrated that diffusion models can produce visually plausible surgical scenes, suggesting the fundamental viability of generative approaches for surgical video synthesis. Having established that diffusion models can generate surgical videos, the next question is how to control what is generated.

\subsection{Conditional Generation}
Conditional generation, leverages conditional diffusion models to synthesize videos guided by explicit conditioning signals. Methods in this category learn conditional distributions $p({V} \mid {C})$, where ${C}$ represents user-specified or task-derived conditions. These signals can take various forms, including texts, phase labels, trajectories, scene graphs, keypoints, kinematic data, or action-related inputs for policy learning and strategy evaluation.
Early efforts introduced basic control modalities. VISAGE \cite{Yeganeh_VISAGE_MICCAI2024W} introduced future video generation in laparoscopic surgery. Given a single initial frame and action graph triplets (instrument, verb, target), the model predicts subsequent frames. An interactive laparoscopic video‑diffusion framework \cite{Iliash_Interactive_DGM4MICCAI2024} enabled interactive generation control via text prompts and instrument segmentation masks. MS-PCD \cite{Zhao_See_MICCAI2024} incorporated procedural planning through phase labels.
SurgSora \cite{Chen_SurgSora_MICCAI2025} proposed object-aware controllable diffusion with dual semantic injection, decoupled flow mapping, and trajectory control. It allows user-guided fine-grained video generation without relying on manually annotated segmentation masks. HieraSurg \cite{Biagini_HieraSurg_MICCAI2025} established a hierarchical two-stage diffusion paradigm, first predicting coarse surgical semantic changes and then generating fine-grained visual features. Ophora \cite{Li_Ophora_MICCAI2025} constructed the first large-scale ophthalmic surgical video-text dataset Ophora-160K and enables text-guided cataract surgery generation. LLaMA-VG \cite{Chen_LLAMAVG_ISBI2025} introduced large vision-language model architectures for endoscopic video generation. SG2VID \cite{SG2VID_MICCAI2025} used scene graphs for fine-grained control of instrument positions and anatomical layouts. MoViS \cite{MoViS_AgenticAI2025} leveraged keypoint-based motion guidance to capture complex instrument kinematics. Blob \cite{Mennillo_Blob_IJCARS2026} explored 3D position-aware surgical scene simulation through kinematic structured modeling.
Conditional generation makes synthesis controllable. However, controllability alone does not constitute understanding how surgical scenes evolve in response to actions.
\subsection{World Modeling Generation}
World modeling generation, aims to generate causally consistent and physically plausible surgical dynamics rather than merely producing visually plausible frames. Unlike conditional generation, which learns $p(\mathcal{V} \mid \mathcal{C})$ from statistical correlations, world modeling generation seeks to capture the underlying transition dynamics $p(s_{t+1} \mid s_t, a_t)$, enabling prediction, interaction, and closed-loop decision-making.
SWoMo \cite{Sivakumar_SWoMo_MICCAI2026} is the representative approach in this category. Its symbolic component combines a rule-based simulator with physical scene graphs to model instrument-tissue interaction dynamics and causal rationality, while a diffusion model renders realistic visual appearance from the predicted states. However, explicit state transition modeling remains an emerging direction with only preliminary demonstrations to date.

\subsection{Summary of Methods}
\begin{table}[ht]
\centering
\caption{Summary of  surgical video generation methods.}
\label{tab:methods}
\resizebox{\linewidth}{!}{
\begin{tabular}{llllc}
\toprule
\textbf{Method} & \textbf{Venue} & \textbf{Paradigm} & \textbf{Control Type} & \textbf{Code} \\
\midrule
Endora \cite{Li_Endora_MICCAI2024} & MICCAI 2024 & Unconditional & None & — \\
VISAGE \cite{Yeganeh_VISAGE_MICCAI2024W} & MICCAI 2024 & Conditional & Image, Structured Label & — \\
Inter. Laparoscopic \cite{Iliash_Interactive_DGM4MICCAI2024} & MICCAI 2024 & Conditional & Text, Mask & — \\
MS-PCD \cite{Zhao_See_MICCAI2024} & MICCAI 2024 & Conditional & Image, Phase Label & — \\
SurgSora \cite{Chen_SurgSora_MICCAI2025} & MICCAI 2025 & Conditional & Image, Trajectory & \href{https://surgsora.github.io}{Link} \\
HieraSurg \cite{Biagini_HieraSurg_MICCAI2025} & MICCAI 2025 & Conditional & Hierarchical Condition & \href{https://diegobiagini.github.io/HieraSurg/}{Link} \\
Ophora \cite{Li_Ophora_MICCAI2025} & MICCAI 2025 & Conditional & Text & \href{https://github.com/mar-cry/Ophora}{Link} \\
LLaMA-VG \cite{Chen_LLAMAVG_ISBI2025} & ISBI 2025 & Conditional & Video, Text & — \\
SG2VID \cite{SG2VID_MICCAI2025} & MICCAI 2025 & Conditional & Scene Graph & \href{https://ssharvienkumar.github.io/SG2VID/}{Link} \\
MoViS \cite{MoViS_AgenticAI2025} & MICCAI 2025 & Conditional & Keypoints & — \\
Blob \cite{Mennillo_Blob_IJCARS2026} & IJCARS 2026 & Conditional & Kinematic Data & — \\
SWoMo \cite{Sivakumar_SWoMo_MICCAI2026} & MICCAI 2026 & World Modeling & Symbolic Action & \href{https://ssharvienkumar.github.io/SWoMo/}{Link} \\
\bottomrule
\end{tabular}
}
\end{table}

\begin{table}[ht]
\centering
\caption{Summary of datasets relevant to surgical video generation.}
\label{tab:datasets}
\resizebox{\linewidth}{!}{
\begin{tabular}{lccccc} 
\toprule
\textbf{Dataset} & \textbf{Modality} & \textbf{Videos} & \textbf{Annotation} & \textbf{Access} \\
\midrule
{Cholec80} & Laparoscopic & 80 & Phase, Tool & \href{http://camma.u-strasbg.fr/datasets/}{Link} \\
{CholecT50} & Laparoscopic & 50 & Phase, Tool, Action & \href{http://camma.u-strasbg.fr/datasets/}{Link} \\
{CoPESD} & Endoscopic ESD & --- & Multi-level Motion & \href{https://github.com/gkw0010/CoPESD}{Link} \\
{Ophora-160K} & Ophthalmic & 9,819 & Text Instructions & \href{https://github.com/mar-cry/Ophora}{Link} \\
{Cataract-1K} & Ophthalmic & 1,000 & Phase, Segmentation & \href{https://github.com/Negin-Ghamsarian/Cataract-1K}{Link} \\
{CATARACTS} & Ophthalmic & 50 & Tool, Activity & \href{https://ieee-dataport.org/open-access/cataracts}{Link} \\
{JIGSAWS} & Robotic & --- & Kinematic, Gesture & \href{https://cirl.lcsr.jhu.edu/research/hardware/datasets/}{Link} \\
{Kvasir-Capsule} & Capsule Endoscopy & 117 & Classification & \href{https://github.com/simula/kvasir-capsule}{Link} \\
\bottomrule

\end{tabular}
}
\end{table}

Table \ref{tab:methods} summarizes representative methods by paradigm, listing publication venue, control type, and code availability where applicable. The three paradigms reflect a progression in generation objective: from unconditional video generation, to controllable conditional generation, to explicit state transition modeling for causally consistent surgical dynamics.

\section{Datasets and Evaluation}\label{Datasets and Evaluation}

\subsection{Public Datasets}
The development of surgical video generation models depends on the availability of high-quality datasets. Table \ref{tab:datasets} summarizes key datasets that have supported recent advances in this field. 


\subsection{Evaluation}
To provide a quantitative reference for the methods surveyed, Table~\ref{tab:experiments} summarizes the reported performance of representative methods on public surgical video datasets. Key metrics include FVD for video quality, FID for frame-level fidelity, and SSIM for structural similarity, with additional metrics such as LPIPS, PSNR, IS, Frame Consistency, and CLIPScore reported where available. It should be noted that direct comparison across datasets is not meaningful due to differences in video length, surgical procedure complexity, and evaluation protocols. 

\begin{table*}[ht]
\centering
\caption{Quantitative results of representative methods on public datasets.}
\label{tab:experiments}
\resizebox{\linewidth}{!}
{
\begin{tabular}{llccccccc}
\toprule
\textbf{Dataset} (\textbf{Setting}) & \textbf{Method} & \textbf{FVD}$\downarrow$ & \textbf{FID}$\downarrow$ & \textbf{SSIM}$\uparrow$ & \textbf{IS}$\uparrow$ & \textbf{PSNR}$\uparrow$ & \textbf{LPIPS}$\downarrow$ & \textbf{Others} \\
\midrule
\multirow{2}{*}{Cholec80 (SG2VID \cite{SG2VID_MICCAI2025})}
& Endora & 533.80 & 47.00  & — & — & — & 0.53 & — \\

& SG2VID & 457.30 & 16.40  & — & — & — & 0.53 & BB IoU: 0.62, F1: 0.47\\

\midrule
\multirow{3}{*}{CholecT50 (MoViS \cite{MoViS_AgenticAI2025})} 

& VISAGE-T & 1780.00 & — & 0.56 & — & 18.10 & 0.39 & — \\
& VISAGE-I & 1875.00 & — & 0.56 & — & 18.30 & 0.38 & — \\
& MoViS & 477.00 & — & 0.57 & — & 18.80 & 0.19 & — \\
\midrule
\multirow{2}{*}{CoPESD (SurgSora \cite{Chen_SurgSora_MICCAI2025})} 
& Endora  & 1146.54   & 205.93 & — & 2.24 & — & — & Frame Cons.: 97.51\% \\
& SurgSora  & 395.65 & 87.94  & 0.56 & 3.28 & 20.71 & — & Frame Cons.: 98.70\% \\
\midrule
\multirow{3}{*}{Cataract-1k (SWoMo \cite{Sivakumar_SWoMo_MICCAI2026})} 
& Endora  & 258.70 & 40.00  & — & — & — & 0.38 & — \\
& SWoMo & 123.00 & 20.10  & — & — & — & 0.39 & BB IoU: 0.65, F1: 0.66 \\
& SG2VID & 73.00 & 24.90 & — & — & — & 0.39& BB IoU: 0.61, F1: 0.62 \\

\hline
\multirow{2}{*}{Cataract-1k (SG2VID \cite{SG2VID_MICCAI2025})} 
& Endora & 265.90 & 30.30  & — & —  & — & 0.38 & — \\
& SG2VID & 901.80 & 137.60 & — & — & — & 0.324 & — \\
\midrule
\multirow{2}{*}{Kvasir-Capsule (LLaMA-VG \cite{Chen_LLAMAVG_ISBI2025})} 
& Endora & 75.95 & 14.03  & — & 2.50 & — & — & — \\

& LLaMA-VG & 59.02 & 10.11  & — & 2.90 & — & — & — \\

\midrule
\multirow{2}{*}{Colonoscopic (LLaMA-VG \cite{Chen_LLAMAVG_ISBI2025})} 
& Endora & 460.70 & 13.41  & — & 3.90 & — & — & — \\
& LLaMA-VG & 369.97 & 12.46  & — & 3.99 & — & — & — \\
\midrule
\multirow{2}{*}{Ophora-160K (Ophora \cite{Li_Ophora_MICCAI2025})} 
& Endora & 990.30 & 60.50  & — & — & — & — & — \\
& Ophora & 276.96 & 33.72  & — & — & — & — & CLIPScore: 39.19 \\

\midrule
\multirow{3}{*}{CATARACTS (SWoMo \cite{Sivakumar_SWoMo_MICCAI2026})} 
& Endora  & 436.80 & 58.40  & — & — & — & 0.46 & — \\

& SG2VID  & 363.80 & 47.30  & — & — & — & 0.44 & BB IoU: 0.50, F1: 0.39 \\

& SWoMo & 265.40 & 40.80  & — & — & — & 0.45 & BB IoU: 0.52, F1: 0.41 \\
\hline
\multirow{2}{*}{CATARACTS (SG2VID \cite{SG2VID_MICCAI2025})} 
& Endora  & 649.50 & 45.90  & — & — & — & 0.46 & — \\

& SG2VID & 523.80 & 40.90  & — & — & — & 0.44 & BB IoU: 0.49, F1: 0.38 \\
\midrule
\multirow{1}{*}{JIGSAWS (Blob \cite{Mennillo_Blob_IJCARS2026})} 
& Blob  & — & — & 0.829  & — & 22.68  & 0.09 & — \\

\bottomrule
\end{tabular}
}

\end{table*}

\section{Limitations and Future Directions}\label{Limitations and Future Directions}
Despite recent progress, surgical video generation faces three key bottlenecks: data and generalization, physical realism and controllability, and evaluation and trustworthiness, which also suggest promising future directions.

\subsection{Data and Generalization}

Most existing methods are trained on datasets from single institutions or specific surgical modalities. Performance under domain shifts across different imaging systems, lighting conditions, or patient populations has not been systematically studied. Unsupervised domain adaptation has been explored in general computer vision to address such distribution shifts \cite{huang2024gradient, huang2026unsupervised}, but whether these techniques transfer to surgical video generation remains unclear.
Data scarcity is another fundamental bottleneck. Most datasets focus on a single procedure and contain few videos, limiting generalization. Surgical video foundation models such as SurgVISTA \cite{SurgVISTA} demonstrate strong transferability across understanding tasks, suggesting that large-scale pre-training may benefit surgical video generation. This direction, however, remains largely unexplored.
Future work should systematically evaluate cross-domain generalization, adapt domain adaptation techniques to surgical video generation, and explore large-scale pre-training on diverse surgical datasets.

\subsection{Physical Realism and Controllability}
Current diffusion models learn visual-temporal distributions from data without explicit modeling of tissue deformation, force transmission, or fluid dynamics. Generated videos may appear visually plausible while violating physical laws, limiting their utility for surgical simulation or preoperative planning.
Conditional generation methods support control via text, trajectory, or phase labels, yet controllability remains limited. Users cannot easily specify complex surgical actions or predict how input changes affect output. More interpretable conditioning mechanisms are needed to enable fine-grained interaction.
Future work should integrate biomechanical priors into generative architectures and develop more interpretable and flexible conditioning mechanisms.

\subsection{Evaluation and Trustworthiness}
Standardized evaluation remains a significant gap. Current practice relies on metrics borrowed from general image and video synthesis, including FVD, FID, IS, PSNR, SSIM, and LPIPS. While these capture visual fidelity, none reflect procedural correctness, anatomical plausibility, or surgical appropriateness. A video may score well on these measures while depicting clinically nonsensical interactions. 
The SurgVeo benchmark \cite{chen2025far}, which introduces the Surgical Plausibility Pyramid (SPP) framework, represents a step toward surgery-specific assessment by evaluating the plausibility of instrument operation, tissue feedback, and surgical intent. However, a community-wide evaluation standard remains absent.
Trustworthiness presents additional concerns. Current models offer little insight into their confidence or failure modes, and uncertainty quantification remains largely unexplored. The internal mechanisms of diffusion models are also difficult to interpret, complicating the assessment of clinical reliability.
Future work should establish community-wide evaluation benchmarks, develop uncertainty quantification methods, and improve interpretability for failure detection and output explanation.

\section{Conclusion}\label{Conclusion}

This survey reviews surgical video generation from 2024 to 2026. We organize the literature into three paradigms according to generation objective: unconditional generation, conditional generation, and world modeling generation. This taxonomy reveals an evolutionary trajectory from establishing synthesis feasibility, to enabling controllability, to modeling causal dynamics for decision-making support.
We summarize publicly available datasets, identify a persistent gap between pixel-level metrics and clinical plausibility, and synthesize key bottlenecks across generalization, physical realism, controllability, and trustworthiness.
World modeling generation offers a promising direction. Achieving it requires explicit state transition modeling for causal consistency and physical plausibility. However, current approaches remain preliminary, and the transition from diffusion to world models is still underway.
Future research should pursue closer integration of domain knowledge, robust evaluation, and physically grounded generation. Progress in these directions is essential for advancing the field toward practical applications.

\bibliographystyle{IEEEtran}
\bibliography{ref}

\end{document}